\def\year{2022}\relax
\documentclass[letterpaper]{article} 
\usepackage{aaai22}  
\usepackage{times}  
\usepackage{helvet}  
\usepackage{courier}  

\usepackage{algorithm} 
\usepackage{algorithmic}  
\usepackage[algo2e]{algorithm2e} 
\usepackage[hyphens]{url}  
\usepackage{graphicx} 
\usepackage{natbib}  
\usepackage{caption} 
\DeclareCaptionStyle{ruled}{labelfont=normalfont,labelsep=colon,strut=off} 
\usepackage{tikz} 
\usepackage{amsmath}
\usepackage{amsfonts}
\usepackage{booktabs}
\usepackage{multirow}

\usepackage{filecontents}
\usepackage{caption}
\usepackage{subcaption}

\usepackage{newfloat}
\usepackage{listings}
\floatstyle{ruled}
\newfloat{listing}{tb}{lst}{}
\floatname{listing}{Listing}

\title{Towards General Deep Leakage in Federated Learning}

\author {
    Jiahui Geng\textsuperscript{\rm 1}\textsuperscript{\textsection}\thanks{Corresponding author},
    Yongli Mou \textsuperscript{\rm 2}\textsuperscript{\textsection},
    Feifei Li \textsuperscript{\rm 2}\textsuperscript{\textsection},
    Qing Li \textsuperscript{\rm 1},
    Oya Beyan \textsuperscript{\rm 3},
    Stefan Decker \textsuperscript{\rm 2},
    Chunming Rong \textsuperscript{\rm 1}
}
\affiliations {
    \textsuperscript{\rm 1} University of Stavanger\\
    \textsuperscript{\rm 2} RWTH-Aachen University\\
    \textsuperscript{\rm 3} University of Cologne\\
    \{jiahui.geng,qing.li,chunming.rong\}@uis.no, \{mou,feifei.li,beyan,decker\}@dbis.rwth-aachen.de, obeyan@uni-koeln.de

}

\usepackage{bibentry}

\begin{document}

\maketitle
\begingroup\renewcommand\thefootnote{\textsection}
\footnotetext{The authors contributed equally to this work.}
\endgroup
\begin{abstract}

Unlike traditional central training, federated learning (FL) improves the performance of the global model by sharing and aggregating local models rather than local data to protect the users' privacy. Although this training approach appears secure, some research has demonstrated that an attacker can still recover private data based on the shared gradient information. 
This on-the-fly reconstruction attack deserves to be studied in depth because it can occur at any stage of training, whether at the beginning or at the end of model training; no relevant dataset is required and no additional models need to be trained.
We break through some unrealistic assumptions and limitations to apply this reconstruction attack in a broader range of scenarios.
We propose methods that can reconstruct the training data from shared gradients or weights, corresponding to the FedSGD and FedAvg usage scenarios, respectively.
We propose a zero-shot approach to restore labels even if there are duplicate labels in the batch. We study the relationship between the label and image restoration. We find that image restoration fails even if there is only one incorrectly inferred label in the batch; we also find that when batch images have the same label, the corresponding image is restored as a fusion of that class of images. Our approaches are evaluated on classic image benchmarks, including CIFAR-10 and ImageNet. The batch size,  image quality, and the adaptability of the label distribution of our approach exceed those of GradInversion, the state-of-the-art.

\end{abstract}

\begin{figure*}
    \centering
    \includegraphics[width=0.9\textwidth]{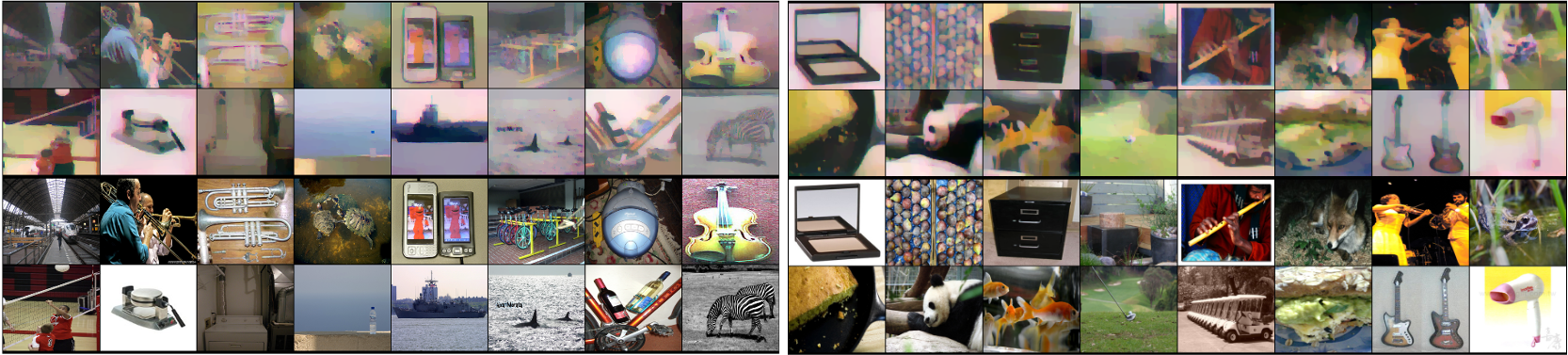}
    \caption{Our Reconstruction Results on ImageNet with Batch Size 16. The left and right pictures show the results of two matches respectively. 8 pictures in a row, the top two rows are recovered pictures, the bottom two rows are ground truth.}
    \label{fig:results16}
\end{figure*}
\section{Introduction}
\noindent With the expansion of information technology and smart devices, massive amounts of data are being produced, collected, stored, processed, and utilized incessantly. 
In the era of big data, it brings broad social changes and security concerns like data abuse and information leakage. 
Society raises the concerns of privacy, which bring the world-widely emergence of data protection legislation, such as the GDPR in the European Union, the HIPAA Act in the USA and the Data Protection Act (PDA) in the UK. 
Central machine learning approaches are becoming insecure due to the potential risk of privacy leakage during data collecting and sharing.

Federated learning (FL), proposed as a privacy-preserving distributed learning paradigm, has gained more researchers' attention. The underlying idea is that all participants collaboratively train a shared model without exchanging local private data.
In the FL scenarios, the central server broadcasts the current model to each client, and each client will train the model on its local data. 
At each communication round, the central server will aggregate gradients or weights updated from a subset of clients into the global model. The corresponding mechanisms were stated in \textit{FedSGD} and \textit{FedAvg} ~\cite{fedavg2016, communication2017}. 

Despite the restricted access to raw data, recent work has shown that FL also faces a variety of privacy attacks, such as membership inference attacks~\cite{exploting2019,DBLP:conf/sp/NasrSH19}, property inference attacks~\cite{exploting2019,DBLP:conf/infocom/WangSZSWQ19}, and reconstruction attacks~\cite{DBLP:conf/ccs/FredriksonJR15,DBLP:conf/ccs/HitajAP17,DBLP:conf/uss/0001B0F020}. Data reconstruction attacks are the most dangerous and challenging attacks with severe consequences. Data reconstruction attacks usually require domain-relevant datasets, additional training of inverse models~\cite{DBLP:conf/ccs/FredriksonJR15}, or GANs~\cite{DBLP:conf/ccs/HitajAP17,DBLP:conf/uss/0001B0F020}.

There has been a series of studies on reconstructing training data based on shared gradients and weights in a FL scenario. These attacks are white-box attacks, where the attacker knows almost all model parameters and hyperparameters, including learning rate and model architecture. Any FL participant that has access to the shared model updates can execute the attack.
DLG \cite{dlg2020} was the first to recover both images and label simultaneously by gradient matching. However, their method is limited by the image scale and batch size.
The authors of iDLG \cite{idlg2020} found that zero-shot label inference was more accurate when batch size equals 1. However, their algorithm did not support the case where the batch size is greater than 1. 
InvertingGradients \cite{invertingg2020} has demonstrated that it is possible to reconstruct images at high resolution (ImageNet) from the gradients no matter trained or untrained models. Total variation dramatically improves the quality of large-scale image recovery. They experimented with images of different labels in minibatch to avoid the ambiguity of the reconstruction. To best of our knowledge, InvertingGradients was the only work that discussed restoration from model weight updates
while their approach can only restore the training data when the batch size or local epochs is equal to 1. 
The latest work GradInversion~\cite{gradinversion2021}  improved InvertingGradients and was feasible when the batch size was up to 48. It also proposed a zero-shot batch label restoration approach, but the approach requires that no images in the batch share the same label and the number of classes must be greater than the batch size.

By analyzing the previous related work, we summarize the limitations as follows:
\begin{itemize}
    \item All the existing approaches did not show the results or could not recover when the batch data had duplicate labels.
    \item All current work independently evaluates image recovery and label recovery tasks and does not discuss the impact of image recovery quality on labels and vice versa.
    \item The previous work on recovering images from weight differences is very restricted.  When the FedAvg algorithm is used in federated learning, the batch size and local epochs will be greater than 1 in almost all cases.
\end{itemize}

In this paper, we try to solve the above-explained problems. Our contributions are as follows:
\begin{itemize}
    \item We empirically point out that the actual range of pixels is meaningful for image restoration. We recommend an image initialization and introduce two regularization terms to improve the image reconstruction.
    \item We propose a zero-shot batch label restoration method that is not constrained by the label distribution.
    \item Better image reconstruction by utilizing multiple updates with regard to the same batch data is implemented.
    \item A framework is designed for reconstructing attacks to recover data from shared gradients and weights in FL scenarios.
    \item We set up the alignment to evaluate the recovered image based on labels and image similarities.
    \item Experimental results of our approaches on various datasets(CIFAR-10 and ImagenNet) are demonstrated. 
\end{itemize}

\section{Preliminaries}
In this section, we introduce the preliminaries related to our work, mainly including the FL mechanism, the threat model, the zero-shot label inference, and other methods to improve the quality of reconstruction, which will help to understand the approaches proposed in Section 3.

\subsection{Federated Learning}
Federated learning is a distributed learning paradigm aiming to preserve the privacy of users. A FL system consists of a parameter server orchestrating the training process and multiple clients holding private local data. A typical federated training process repeats the five steps: client selection, broadcast, client computation, aggregation, and model update~\cite{advanced2019}. The server selects a subset of clients and broadcasts the global model at the communication round. Then the clients execute local training on its private data. Eventually the server aggregates the updated models from clients and updates the global model. Hence, the local privacy is preserved. 

Assuming that the shared model architecture is $F$, in the $t$-th communication round the global model parameter is $W^t$, the local model is $W^t_i$, on the $i$-th node,  the training data is ($x^{t_i}$, $y^t_i$), the local client will calculate the local update $ \nabla W^t_i$ as 
\begin{equation}
        \nabla W^t_i = \frac{\partial \mathcal{L}(F(x^t_i, W^t), y^t_i)}{\partial W^t} .
    \end{equation}       

The global model update based on gradient aggregation and  weight aggregation are as follows:
\begin{itemize}
    \item FedSGD
    \begin{equation}
        \nabla W^t = \frac{1}{N} \sum_i^N \nabla W^t_i
    \end{equation}
    
    \begin{equation}
        W^{t+1}= W^t - \eta \nabla W^t,
    \end{equation}
    \item FedAvg
    \begin{equation}
        W^{t+1}_i = W^t -\eta \nabla W^t_i
    \end{equation}
    \begin{equation}
        W^{t+1}= \frac{1}{N} \sum_i^N W^{t+1}_i.
    \end{equation}    
\end{itemize}
Here we show the generic formulas. More details like the weights of each clients, and the  hyperparameters to balance model accuracy and communication efficiency are not reflected above.

\subsection{Threat Model}
\paragraph{Attacker's Goals and Capabilities}
In this work, we study a data reconstruction attack. It's a white-box attack that the attacker knows model architecture, optimizer, loss function, hyperparameters like learning rate, batch size, etc. 
Corresponding to the two FL algorithms FedSGD and FedAvg, we define two attack scenarios respectively, (1) the attacker has access to model weights $W^t_i$ and gradients $\nabla W^t_i$ at the communication round $t$ in case FedSGD; and (2) the attacker has access to weights $W^t_i$ and $W^{t+1}_i$ at communication rounds $t$ and $t+1$ in case FedAvg.
The attacker aims to reconstruct the batch data (images and labels) based on shared model weights or gradients. 
The principle is different from other reconstruction attacks like Model Inversion~\cite{DBLP:conf/ccs/FredriksonJR15} or GAN attacks~\cite{DBLP:conf/ccs/HitajAP17,DBLP:conf/uss/0001B0F020}, which requires the loss of the learning system or image representation, a related public dataset and training of extra additional models.

\paragraph{Data Restoration by Gradient Matching}
Deep Leakage from Gradients(DLG) algorithm was the first implementation to prove that it is possible to obtain private training data from shared gradients. It made the dummy input close to the original input by minimizing the difference of the original gradient and dummy gradient. 
First, dummy data and labels are randomly generated.
After deriving the dummy gradient w.r.t. the dummy data and dummy label, it did not update the model weights like traditional learning processes but updated the dummy data and label to restore privacy by matching the dummy gradient and the original gradient. The DLG algorithm flow is illustrated in Algorithm~\ref{algo:dlg}:
\begin{algorithm}

\caption{Deep Leakage from Gradients}
\SetAlgoLined
\KwIn{$F(x; W)$: differential model; $W$: differentialable model weights; $\nabla W$ gradients; $N$: number of iterations}
\KwOut{$\hat{x}_{N+1}$: restored image; $\hat{y}_{N+1}$: restored label} 
\BlankLine
Initialize the dummy image and label $\hat{x}_1 \hookrightarrow  \mathcal{N}(0, 1)$, $\hat{y}_1 \hookrightarrow  \mathcal{N}(0, 1) $\\
\For{\textnormal{i = 1,...,N}}{
    $\hat{\nabla W_i} = \frac{\partial \mathcal{L}(F(\hat{x}_i, W), \hat{y}_i)}{\partial W} $\\
    ${\mathbb{D}_i = \left\Vert \hat{\nabla W_i} - \nabla W \right\Vert^2 }$\\
    $\hat{x}_{i+1} = \hat{x}_i - \eta \nabla_{\hat{x}_i} \mathbb{D}_i, \quad \hat{y}_{i+1} = \hat{y}_i - \eta \nabla_{\hat{y}_i} \mathbb{D}_i$
}
\label{algo:dlg}
\end{algorithm}

\subsection{Zero-shot Label Restoration}
Zero-shot label restoration was first introduced in iDLG~\cite{idlg2020}. The zero-shot method is popular ~\cite{invertingg2020,gradinversion2021} because the attack is fast and doesn't depend on the image recovery, especially when the input, or model, is too complex. 
We present here the principle of this zero-shot approach, and in Section~\ref{sec:label-restoration} we will describe how to make this approach support batch restoration for any label distribution.

Assume the one hot label is $y$ and $y_c=1$ for ground truth $c$, the final logits output before the softmax is $z$,  the number of classes is $N$, so prediction $p_i = \frac{e^{z_i}}{\sum_{n=1}^Ne^{z_n}}$.
The cross-entropy loss for given training data $(x,y)$ is 
\begin{equation}
   \mathcal{L}(x,y) = - \sum_{n=1}^N y_n \log{\frac{e^{z_n}}{\sum_{n=1}^N e^{z_n}}} = - \log{\frac{e^{z_c}}{\sum_{n=1}^N e^{z_n}}}
\end{equation}
Then the gradient of the loss w.r.t the logit at index $i$ is 
\begin{equation}
    \frac{\partial \mathcal{L}(x,y)}{\partial z_i} = - \frac{\partial z_c - \log \sum_{n=1}^N {e^{z_n}}}{\partial z_i} = p_i - y_i,
    \label{eq:logit-grad}
\end{equation}
if $i=c$, $y_i=1$, else $y_i=0$. 

\subsection{Total Variation Denoising} A commonly used noise reduction method in signal processing is total variation~\cite{DBLP:conf/cvpr/MahendranV15}. 
Given a signal $x$, the goal of total variation denoising is to find an approximation signal $y^{*}$ such that it has smaller total variation but is also "close" to image $x$ as the objective function $y^{*} = \arg\min [ E(x, y) + TV(y) ]$.

Considering the 2D image restoration task, a restored image containing noise will have a larger value of total variation, i.e., a larger sum of absolute values of its gradients, compared to its signal unaffected by noise. 
\begin{equation}
\label{eq:tv-aniso}
    TV(X) =  \sum_{i,j} (|X_{i+1,j} - X_{i,j}| + |X_{i, j+1} - X_{i, j}|)  
\end{equation}

\subsection{Image Similarity} In our work, image similarity is not only used as the metric of image restoration but also for image alignment. When there are multiple images in the batch with the same label, we need to align the images according to the image similarity. Previous works~\cite{dlg2020,invertingg2020,gradinversion2021} have confirmed that gradient matching-based recovery loses the order of batch images, and the ordered results are useful for analyzing the experimental effects. For images $x$ and $y$, there are several methods to measure the similarity:
\begin{itemize}
    \item Mean square error (MSE)
    \[ MSE(x, y) = E{(x-y)^2},\]
    where $E$ is the expected value.
    \item Peak signal to noise ratio (PSNR)
    \[PSNR = 10 * \log_{10}(\frac{MAX^2}{MSE(x, y)}), \]
    where $MAX$ is the maximum possible pixel value of the image, for $n$-bit pixel, $MAX=2^n-1$. Higher means more similar.
    \item Structural similarity (SSIM)
    \[ SSIM(x, y) = \frac{(2\mu_x\mu_y+c_1)(2\sigma_{xy}+c_2)}{(\mu_x^2+\mu_y^2+c_1)(\sigma_x^2+\sigma_y^2+c_2)}, \]
    where $\mu$, $\sigma$ are the corresponding mean and standard variation; $\sigma_{xy}$ is the covariance between images; $c_1=(k_1L)^2$, $c_2=(k_2L)^2$ are used to stabilize the division with $k_1=0.01$, $k_2=0.03$. Higher means more similar.
    \item Learned perceptual image patch similarity (LPIPS)
    
    evaluate the distance between image pairs with deep learning models, higher value means further difference.
\end{itemize}

\section{Proposed Approaches}
In this section, we explain our proposed method in detail. In Section~\ref{sec:rec-func} we introduce the reconstruction function, explaining how the reconstruction attack is defined as an optimization problem. In Section~\ref{sec:label-restoration} we introduce our zero-shot batch label restoration approach. In Section~\ref{sec:auxiliary} we introduce our auxiliary regularization designed to improve the fidelity of image reconstruction. In Section~\ref{sec:multi-updates} we explain how to exploit multiple updates, which further improve the reconstruction results. In Section~\ref{sec:weights} we explain how to apply our reconstruction attack architecture to the FedAvg algorithm, that is, to reconstruct the training images according to their weights. In Section~\ref{sec:align} we describe how to implement image alignment, which helps us to analyze the data and eliminate ambiguity between images.

\subsection{Reconstruction Function}
\label{sec:rec-func}
The original private data is $(x,y)$, the recovered private data is $(\hat{x},\hat{y})$. The parameters of the model is $W$ and $\nabla W$ is the batch averaged gradient in the FedSGD scenario. 
The reconstruction function is illustrated as Equation~\ref{eq:dlg-objective-gradients}, consisting of gradient matching loss and auxiliary regularization for fidelity:

\begin{equation}
    \hat{x}^*, \hat{y}^* = \arg\min_{\hat{x},\hat{y}} {[ \mathcal{L}_{gm}(\hat{x},\hat{y}, W, \nabla W) + \mathcal{R}_{aux}(\hat{x})]}.
    \label{eq:dlg-objective-gradients}
\end{equation}

Gradient matching loss describes the discrepancy between the original and the dummy gradient among all layers. Hence the first part is defined in Equation~\ref{eq:gradient} and $l$ denotes the model layer:
\begin{equation}
    \mathcal{L}_{gm}(\hat{x}, W, \nabla W) = \sum_{l} \left\Vert \nabla_{W_{(l)}} \mathcal{L}(\hat{x}, \hat{y}) -\nabla W_{(l)} \right\Vert_{2}.
    \label{eq:gradient}
\end{equation}

Auxiliary regularization for fidelity will be explained in detail in Section~\ref{sec:auxiliary}.

\subsection{One-shot Batch Label Restoration}
\label{sec:label-restoration}
Here we explain our zero-shot approach, which solves $\hat{y}^*$ according to the gradient of the fully connected layer. We consider the information leakage from the classification task, in which cross-entropy is used as loss function for optimization. 
We denote $x^{*} = [x_{1}, x_{2}, \cdots, x_{K}]$ and $y^{*} = [y_{1}, y_{2}, \cdots, y_{K}]$ as the batch of original images and ground truth labels, respectively.
For each image-label pair $(x_{k}$, $y_{k})$ in the batch, we calculate the cross entropy as Equation~\ref{eq:cross-entropy}, where $z_{k,n}$ is the logits output of the last fully connected layer and $y_{k, n} \in \{0, 1\}$ is the one-hot representation of $y_{k}$ at index $n$ among all $N$ classes.
\begin{equation}
    \mathcal{L}(x_{k}, y_{k}) = -\sum_{n=1}^{N} y_{k,n} \log(\frac{e^{z_{k,n}}}{\sum_{n=1}^{N} e^{z_{k,n}}})
    \label{eq:cross-entropy}
\end{equation}

The gradient of loss function w.r.t. weights $W$ is formulated as Equation~\ref{eq:gradient_def}, where $K$ is the number of batch size.
The gradient of cross entropy loss w.r.t. the final logits $z$ at index $n$ of sample $k$ in the batch is derived as Equation~\ref{eq:gd-l-z}, where $p_{k, n} = \frac{e^{z_{k,n}}}{\sum_{n'=1}^{N} e^{z_{k,n'}}}$ is the post-softmax probability. 

\begin{equation}
    \nabla_{W}\mathcal{L}(x^{*}, y^{*}) = \frac{1}{K} \sum_{k=1}^{K}\nabla_{W}\mathcal{L}(x_{k}, y_{k})
    \label{eq:gradient_def}
\end{equation}

\begin{equation}
    \nabla_{z_{k,n}} \mathcal{L}(x_{k}, y_{k})  = p_{k,n} - y_{k,n}
    \label{eq:gd-l-z}
\end{equation}

We define $W^{(FC)} \in \mathbb{R}^{M \times N}$ as the weights of last fully connected layer in the network, where $M$ is the dimension of last hidden layer. 
$\Delta {W^{(FC)}_{m,n,k}} := \nabla_{W^{(FC)}_{m,n}} \mathcal{L}(x_k, y_k)$ are defined as the gradient of the cross-entropy loss for image-label pair $(x_k, y_k)$ w.r.t. weights $W^{(FC)}_{m,n}$ connecting $m$-th unit of the last hidden layer to logits $n$. 
Equation~\ref{eq:dW-m-n-k} is derived according to the chain rule, where $O_{k,m}$ is the $m$-th input unit of fully connection layer of $k$-th sample.
We denote $\Delta W^{(FC)}_{m,n}$ as the average of tensor $\Delta W^{(FC)}$ along the dimension of batch index $k$, i.e., $\Delta W^{(FC)}_{m,n} = \frac{1}{K}\sum_{k=1}^{K} \Delta W^{(FC)}_{m,n,k}$, where

\begin{equation}
\begin{split}
    \Delta W^{(FC)}_{m,n,k} &=  \frac{\partial \mathcal{L}(x_{k}, y_{k})}{\partial z_{k,n}} \cdot \frac{\partial z_{k,n}}{\partial W^{(FC)}_{m,n}}  \\
   & = \nabla_{z_{k,n}} \mathcal{L}(x_{k}, y_{k}) \cdot O_{k,m} \\
   & = (p_{k.n} - y_{k,n}) \cdot O_{k,m}
    \label{eq:dW-m-n-k}
\end{split}
\end{equation}

We denote $S_{k,n}$ as the sum of tensor $\Delta W^{(FC)}$ along the dimension of input features of fully connected layer $m$, i.e., $S_{k,n} = \sum_{m=1}^{M} \Delta W^{(FC)}_{m,n,k}$.
Note that the common activation functions, like ReLU and Sigmoid, in the Convolutional Neural Networks (CNNs), the $O_{k,m}$ are non-negative. Therefore, the sign of $\Delta W^{(FC)}_{m,n,k}$ is only determined by $p_{k,n} - y_{k,n}$, and only one element of $S_{k,n}$ along the dimension of batch index $k$ is negative, if and only if $n = n^{*}_{k}$ at the index of ground truth label. 

The sum of values in $\Delta W^{(FC)}$ along $m$ dimension is denoted as $\Delta  W^{(FC)}_n$:
\begin{equation}
\begin{split}
    \Delta W^{(FC)}_n & = \frac{1}{K}\sum_k \sum_m{(p_{k,n} - y_{k,n}) \cdot O_{k,m}} \\
    & = \frac{1}{K} \sum_k (p_{k,n} - y_{k,n}) \cdot \sum_m O_{k,m} 
\end{split}
\end{equation}
We notice that $O_k = \sum_m O_{k,m}$ is always greater than 0 and depends on each sample. Assume that $O_k\approx \Bar{O_k}$, close to is mean value, then $\Delta W^{(FC)}_n \approx \frac{1}{K}\sum_k {(p_{k,n} - y_{k,n})}\Bar{O_k}$. 
Based on that assumption, we can infer the number of specific labels $n$ by
\begin{equation}
    \label{eq:dblr}
   \sum_k {y_{k,n}} \approx \sum_k {p_{k,n}}-\frac{K\Delta W^{(FC)}_n}{\Bar{O_k}} ,
\end{equation}
the left part implies the number of samples where labels equal $n$ on one batch and we estimate $p_{k,n}$ and $\Bar{O_k}$ by feeding the dummy data into the network. We experimentally demonstrate the validity of our hypotheses and approaches. 
\subsection{Auxiliary Regularization for Fidelity}
\label{sec:auxiliary}
A significant difference between data recovery and standard deep model training is that the pixel value has a physical meaning, usually between $0$ and $255$, or after normalization $0 \sim 1$. 
In contrast, the recovered data from deep leakage are not bounded. Therefore, we must constrain the values of pixels. We have designed two regularization terms as Equation ~\ref{eq:clip-reg},\ref{eq:scale-reg} for this purpose:

\begin{equation}
     \mathcal{R}_{clip}{(\hat{x})} = \left\Vert {\hat{x} - clip(\hat{x})} \right\Vert_2
    \label{eq:clip-reg}
\end{equation}

\begin{equation}
     \mathcal{R}_{scale}{(\hat{x})} = \left\Vert {\hat{x} - scale(\hat{x})} \right\Vert_2
    \label{eq:scale-reg}
\end{equation}
where $clip(\hat{x}_i) = min(max(\hat{x}_i, 0), 1)$ and $scale(\hat{x}_i) =  \frac{\hat{x}_i - min_v}{max_v -  min_v}$. $max_v$, $min_v$ denote the maximum and minimum value in the restored image. The auxiliary regularization (Equation~\ref{eq:aux-reg}) keeps the recovered images away from unrealistic images. $\mathcal{R}_{aux}(\hat{x})$ is the total variation loss. 
\begin{equation}
    \mathcal{R}_{aux}(\hat{x}) =  \alpha_{tv} \mathcal{R}_{tv}(\hat{x})  +  \alpha_{c} \mathcal{R}_{clip}(\hat{x})  +  \alpha_{s} \mathcal{R}_{scale}(\hat{x})
     \label{eq:aux-reg}
\end{equation}

In addition, we found that constraining the initialized pixel between [0,1) also helps to improve the quality of the recovered image as illustrated in our ablation studies.

\subsection{Multiple Updates Consistency}
\label{sec:multi-updates}
\begin{algorithm}
\caption{Recovery from Multiple Updates}
\SetAlgoLined
\KwIn{$\{F_{j}(x; W)\}_{j=1}^{M}$: differential model; $T$: total update times;  $\{W^t\}, t = 1..T$: model weights at different timestamps; $\{\nabla W^t\}, t= 1..T $ corresponding gradients;N: number of iterations.}
\KwOut{$\hat{x}_{T+1}$: private image; $\hat{y}$: private label} 
\BlankLine
Initialize the dummy image $x \hookrightarrow  \mathcal{U}(0, 1)$, recovered label $\hat{y}$ using zero-shot approach.\\
\For{\textnormal{i = 1,...,N}}{
    \For{\textnormal{t = 1,...,T}}{
        $\hat{\nabla W^t_i} = \frac{\partial \mathcal{L}(F(x_i, W^t), \hat{y})}{\partial W^t} $\\
    }
    ${\mathbb{D}_i = \frac{1}{T} \sum_t \left\Vert \hat{\nabla W^t_i} - \nabla W^{t} \right\Vert^2 }$\\
    $x'_{i+1} = x'_{i} - \eta \nabla_{x'_i} \mathbb{D}_i$
}
\end{algorithm}
Just like finding the numerical solution to an equation, when more conditions are known, the more accurate the resulting solution will be. Inspired by this idea, we can obtain better images by increasing the number of known weight and gradient pairs. 
Supposed that one FL participant has trained models locally on the same private data and update the models and gradients for multiple times $T$. By matching the gradients of several different timestamps, ${\frac{1}{T} \sum_t \left\Vert \nabla W'_{i,t} - \nabla W_{t} \right\Vert^2 }$, we can hopefully recover a better batch of images.
It can be considered as a consistency regularization that helps get closer to the global optimal point.

\subsection{Reconstruction from Weights}
\label{sec:weights}
 For model training with SGD optimizer, the model updates as follows:
\begin{equation}
     W^{t+1} = W^{t} - \eta \nabla W^{t}
\end{equation}
Suppose the model undergoes $T$ time of local updates, from $W^0$ to $W^T$, and is finally uploaded the server. That is, the server can access $W^0$ and $W^T$, but not the intermediate models. Our reconstruction attack based on weight updates can be expressed as:
\begin{equation}
     \hat{x}^*,\hat{y}^* = \arg\min_{\hat{x},\hat{y}} {[ \mathcal{L}_{gm}(\hat{x},\hat{y}, W^{0}, W^{T}) + \mathcal{R}_{aux}(\hat{x})]}
     \label{eq:dlg-objective-weights}
\end{equation}
As in common settings of FL, the center server knows all hyperparameters about local training, such as local epochs, local batch size, and local learning rate. Theoretically, we can infer the gradient w.r.t. the training data from continuous model weight updates with $\nabla W^{t+1} = \frac{1}{\eta} {(W^t - W^{t+1})}$, assuming that the data hold on specific clients remains consistent. When T is greater than 1, we adopt a simple strategy by assuming that:
\begin{equation}
    W^T\approx W^0 - T \cdot \eta \nabla W^0
\end{equation}
\begin{equation}
    \nabla W^0 \approx \frac{1}{T\eta}( W^{0} - W^{T}).
\end{equation}
That is, we use the average model difference, denoted as fake gradient, to replace the gradient in the original reconstruction function. We empirically demonstrate that even with the fake gradient, our proposed methods can still restore the image and label information. Gradient matching loss based on weights is defined in Equation~\ref{eq:weight}, where we use cosine similarity instead of $l2$:

\begin{equation}
    \sum_l \left\Vert \nabla_{W^0_{(l)}} \mathcal{L}(\hat{x}, \hat{y}) - (\frac{W_{(l)}^{0} - W_{(l)}^{T}}{T\eta}) \right\Vert_{cos}
    \label{eq:weight}
\end{equation}
\subsection{Image Alignment}
\label{sec:align}

Because the reconstruction based on gradient matching cannot restore the sequence~\cite{dlg2020}, and the same label in the mini-batch causes the ordering ambiguity w.r.t that label~\cite{invertingg2020}, all the image restoration results are not aligned with the ground truth or rely on manual processing. We propose that pictures can be aligned based on the similarity of the pictures. We consider the alignment as a weighted bipartite graph problem. 

Suppose $K$ is the batch size, and a matching function $m$, $\forall i \in 1, \hdots,N$,  $m(i) \in 1, \hdots, N$, needs to be found. The mapping is bijective, $\forall i, j \in 1, \hdots, K$, $i \neq j \Rightarrow m(i) \neq m(j) $.
$sim(\hat{x}_i, x_{m(i)})$ denotes the image similarity between the restored image $\hat{x}_i$ and its aligned ground truth $x_{m(i)}$.
The objective is to maximizing the sum of the similarity in the perfect matching:
\begin{equation}
    m^* = \arg\max_m \sum_{i=1}^N sim(\hat{x}_i, x_{m(i)})
\end{equation}

We use label information to help determine partially aligned images. If there exist labels, which are unique both in the original and restored labels, the images are aligned directly. These aligned images will not be involved in the optimization, which reduces the number of interfering items in the image matching process. We use PSNR as the similarity function when aligning images because it has enough distinction and requires no additional processing.
\begin{figure}
    \centering
    \includegraphics[width=0.4\textwidth]{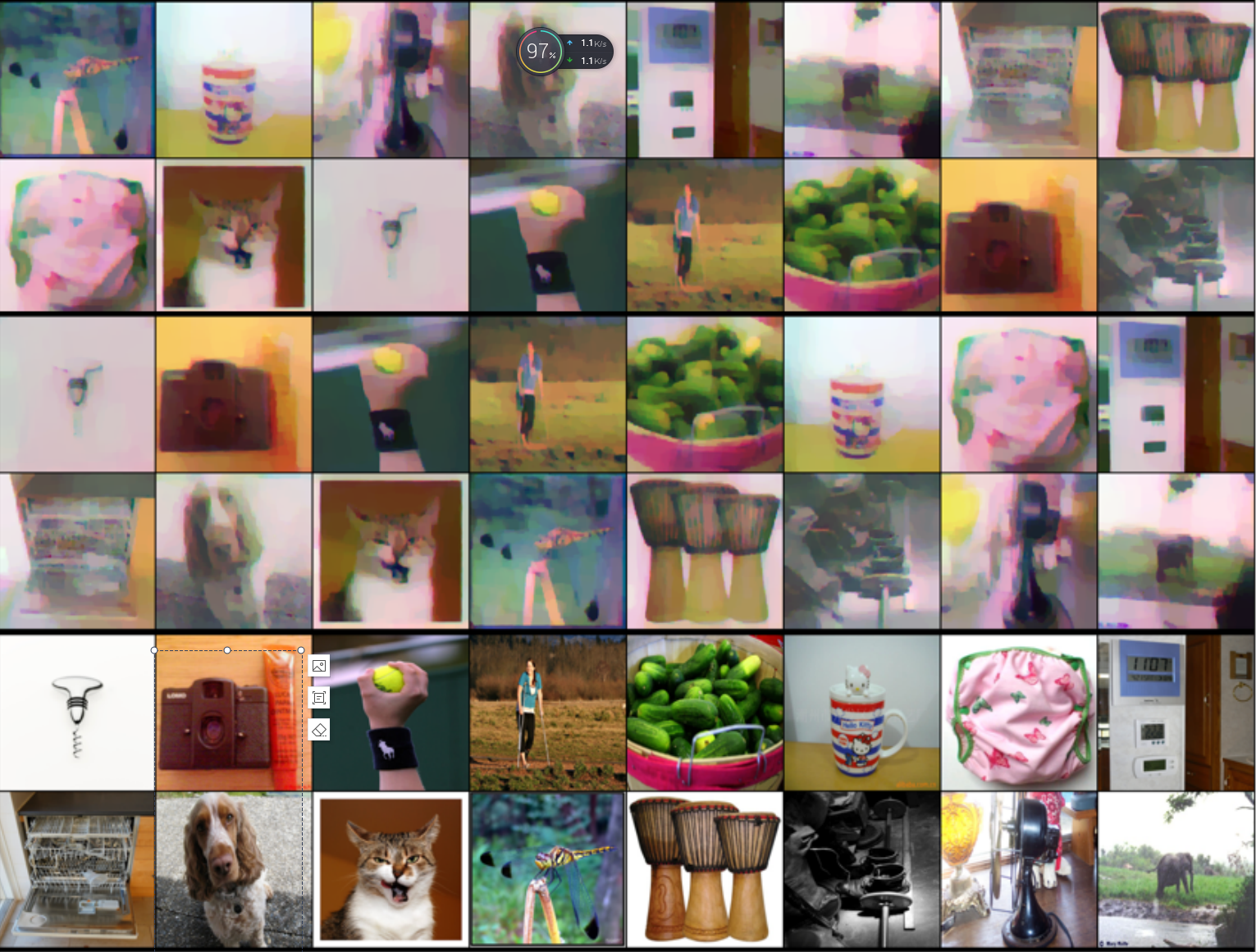}
    \caption{Alignment of unique images from ImageNet of batch size 16. The top two lines are the result of restoration, the middle two lines are the result of image alignment, and the last two lines are ground truth.}
    \label{fig:mix1}
\end{figure}

\section{Experiments}
\paragraph{Setups}
We perform data reconstruction attacks against LeNet networks in FedSGD and FedAvg scenarios. We evaluate our method for the image classification on two classic datasets: CIFAR-10 and ImageNet.  In CIFAR-10, images are with size of $32\times32$ and ImageNet images are cropped to $224\times224$. We sample 10K images from CIFAR-10, 1K for each class and 1K images from ImageNet, five for each of 200 categories.
We use SGD as the optimizer for local training with learning rate(lr) equals $0.1$. We use L-BFGS optimizer and optimize for 300 and 600 iterations respectively for CIFAR-10 and ImageNet. For CIFAR-10 experiments, we set $lr = 0.5$, $\alpha_{tv} = 0.5$, $\alpha_{l2}=0.1$,  $\alpha_{c}=100$, and $\alpha_{s}=100$; For ImageNet we set $lr = 0.1$, $\alpha_{tv} = 200$, $\alpha_{l2}=200$,  $\alpha_{c}=200$, and $\alpha_{s}=200$. We synthesize image batches with NVIDIA V100 GPUs. 
Our experimental results are mainly compared with GradInversion, the start-of-the-art.

\paragraph{Comprehensive Results} Table~\ref{tab:summary} shows the similarities between the images we recovered and ground truth in ImageNet (batch size=16). All labels in the batch are unique. Our method outperforms GradInversion in restoring batch images. The restoration results are better when multiple updates are exploited. The images of our restoration are in Figure~\ref{fig:results16}.
\begin{table}
\centering
\caption{Comparison of GradInversion and our approaches on unique batch of size 16 from ImageNet. }
\label{tab:summary}
\resizebox{\columnwidth}{!}{
\begin{tabular}{cccccc}
\toprule
       & MSE $\downarrow$      & SSIM $\downarrow$    & PSNR $\uparrow$ & LPIPS $\downarrow$    \\ 
\midrule
GradInversion           & 0.015  &  0.701 &19.956 & 0.350      \\ \hline
Ours (Baseline)         & 0.011   &  0.705 & 20.029   & 0.349   \\ \hline
Ours (2 Updates)         & 0.008   &  0.775 & 22.436   &  0.321    \\ \hline
Ours (3 Updates)    & 0.008  &  0.782 & 23.491  &  0.313    \\ \hline
Ours (4 Updates)    & 0.006  &  0.801 & 23.240   &  0.305     \\ \hline
\end{tabular}%
}
\end{table}

\begin{figure}
    \centering
    \includegraphics[width=0.525\textwidth]{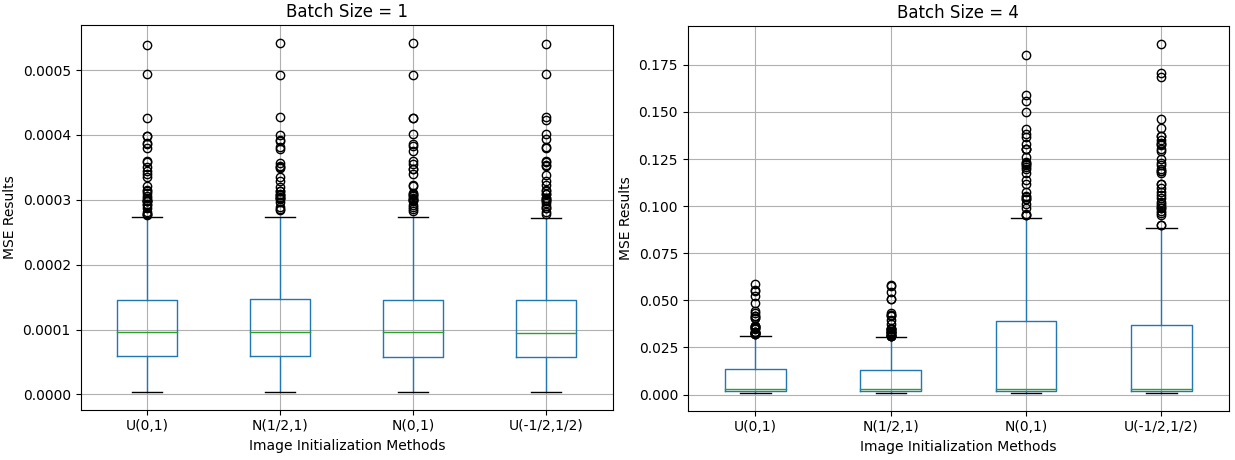}
    \caption{Comparison of the four initialization methods with batch size 1 and 4.}
    \label{fig:init}
\end{figure}

\paragraph{Ablation studies of image initialization and regularization}
We adopt four kinds of random initializing distributions, as $\mathcal{U}(0, 1)$, $\mathcal{N}(0.5, 1)$, $\mathcal{N}(0, 1)$, $\mathcal{U}(-0.5, 0.5)$. 
The results illustrated in Figure~\ref{fig:init} show the influence of those four initialization strategies.We find that when the batch size increases, the initialization of the image with an initial mean value of 0.5 ($\mathcal{N}(0, 1)$,  $\mathcal{U}(0, 1)$) is better than the initialization of the image with a mean value of 0 ($\mathcal{N}(0.5, 1)$, $\mathcal{U}(-0.5, 0.5)$). The ablation study of regularization methods in Table~\ref{tab:ablation} shows that regularization we designed helps to improve the quality of the images

\begin{table}
\caption{Ablation Study of Different Regularization Methods}
\label{tab:ablation}
\resizebox{\columnwidth}{!}{
\begin{tabular}{ccccc}
\toprule
Regularization Methods   & MSE $\downarrow$ & SSIM $\uparrow$ & PSNR $\uparrow$ & LPIPS $\downarrow$\\ 
\midrule
+tv                 & 0.012021    & 0.789463    & 22.090446     & 0.040423      \\ \hline
+tv+l2              & 0.123710    & 0.786881    & 21.967402     & 0.041161      \\ \hline
+tv+clip            & 0.008571    & 0.823602    & 22.723407     & 0.032694      \\ \hline
+tv+scale           & 0.008571    & 0.823608    & 22.723688     & 0.032691      \\ \hline
+tv+clip+scale      & 0.008570    & 0.823609    & 22.723748     & 0.032692        \\ \hline
+tv+l2+clip+scale   & 0.008585    & 0.823514    & 22.715970     & 0.032721      \\
\bottomrule
\end{tabular}
}
\end{table}

\paragraph{Label Inference}
We compare our approach to GradInversion's zero-shot label restoration. 
Tabel~\ref{tab:label-repeats} shows the label accuracy on ImageNet when the batch size is equal to 16. \textit{Repetition} represents the number of extra repetitions. When \textit{Repetition} equals 8, one label appears 9 times, 8 are the extra repetitions. GradInversion can only restore one of the repeated labels. When the \textit{Repetition} increases, the label accuracy decreases linearly. When repetition is equal to 8, the 8 extra repetitions cannot be inferred correctly and the accuracy rate is close to 50\%, actually 49.39\%. Our approach is not affected by the number of duplicate labels.  

The results in Table~\ref{tab:label-our} demonstrate that our approach is robust to duplicate labels, very large batch size and number of classes. \textit{Rep Amount} denotes the number of occurrences of each label in the batch. When \textit{Rep Amount} equals 4 and the batch size equals 256, there are 64 unique labels.

\begin{table}
\caption{The comparison between our method and GradInversion on the label recovery accuracy(\%) of ImageNet when batch size = 16. }
\centering
\resizebox{\columnwidth}{!}{
\begin{tabular}{cccccccccc}
\toprule
  \textit{Repetition}       & 0 & 1  & 2 & 3 & 4 & 5 & 6 & 7 & 8   \\
\midrule
GradInversion &   99.60 & 92.47 & 87.10 & 81.05 & 74.60 & 68.15 & 62.30 & 55.65 & 49.39     \\ 
Our &             99.19 & 98.99 & 97.38 & 98.20 & 98.19 & 99.19 & 99.193 & 98.19 & 99.60   \\ 
\bottomrule
\end{tabular}%
}

\label{tab:label-repeats}
\end{table} 

\begin{table}
\caption{Label accuracy(\%) of our approach under different batch sizes and different repeat amounts.}

\resizebox{\columnwidth}{!}{
\begin{tabular}{ccccccccc}
\toprule
Dataset &\textit{Rep Amount}\textbackslash{}\textit{Batch Size} & 4 & 8 & 16 & 32 & 64 & 128 & 256 \\ 
\midrule
\multirow{2}{*}{CIFAR-10}       & 2      &  100 & 99.63 & 98.06 & 97.13 & 98.03 & 97.97 & 97.83  \\
                        & 4   &  99.30 & 98.60 & 99.50& 98.48 & 98.75 & 94.87 & 98.44  \\
\midrule
\multirow{2}{*}{ImageNet}       & 2      & 99.00 & 100 & 98.39 & 97.50 & 98.44 & 94.20 & 96.35 \\ 
                        & 4      &  99.30 & 98.60 & 99.50& 98.48 & 98.75 & 94.87 & 98.44 \\ 
\bottomrule
\end{tabular}
}
\label{tab:label-our}
\end{table}

\begin{figure}
    \centering
    \includegraphics[width=0.4\textwidth]{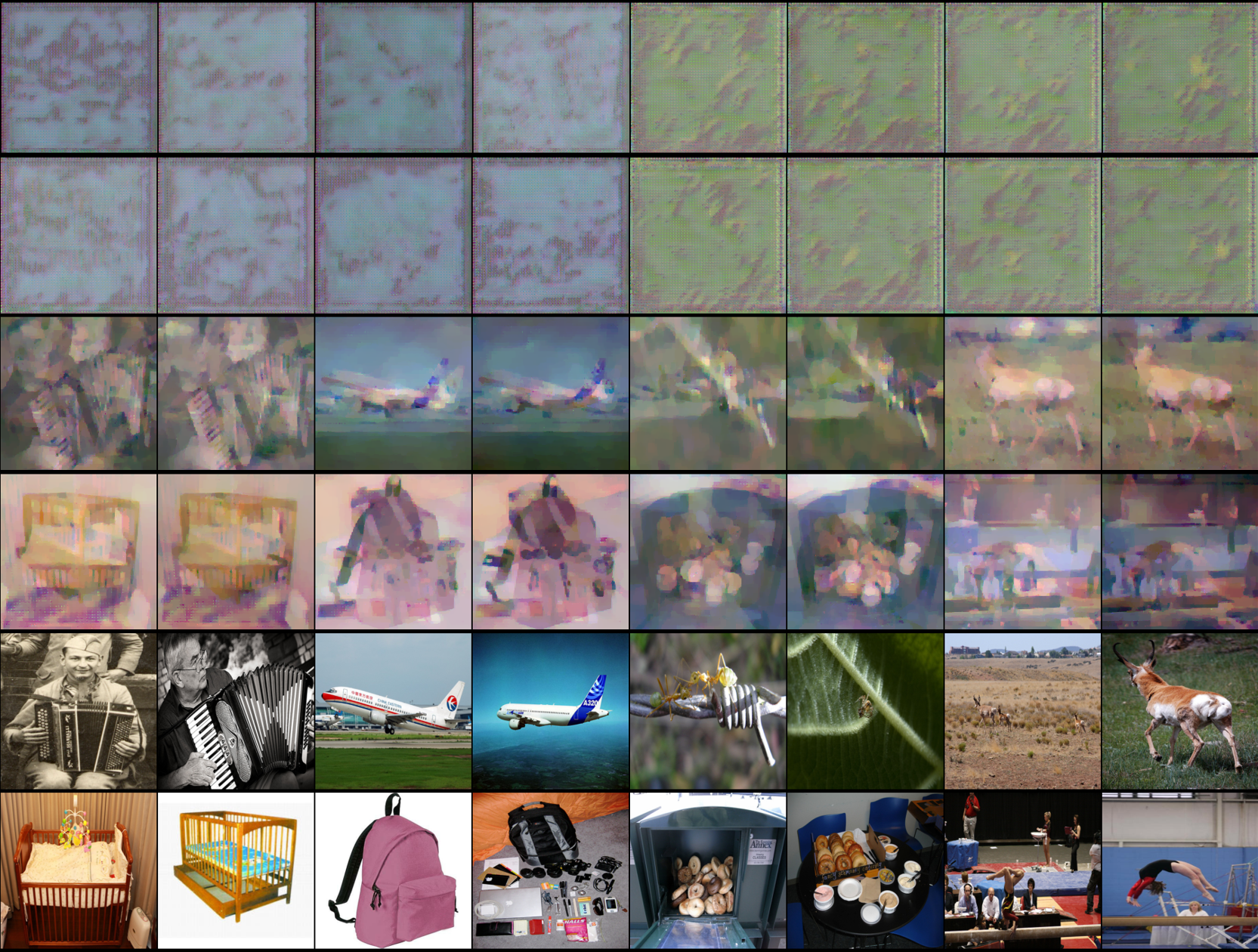}
    \caption{Batch image reconstruction on ImageNet dataset using different approaches with \textit{Rep Amount} 2. }
    \label{fig:compare}
\end{figure}

\paragraph{Impact of Labels on Image Restoration}
Table~\ref{tab:compare} shows the results of image restoration when there are repeating labels in one batch. \textit{Reptition} and \textit{Rep Amount} are the settings explained above.
Each cell in the table contains two values, the first value is for GradInversion and the second value is for ours. 
GradInversion can not handle duplicate labels correctly, so even when \textit{Repetition} equals 1, only one image is incorrectly labeled, the discrepancy between the restored image and the ground truth gets extremely large (MSE: 0.199 , PSNR: 10.639). In contrast, our method is able to infer the duplicate labels properly, and the image recovery is minimally affected(MSE: 0.018, PSNR:19.122). In addition, we found that images with the same label are recovered as a fusion of images in our approach, although not identical, but the outlines of similar images can be seen in the images. Figure~\ref{fig:compare} shows the result of image recovery when Rep Amount=2 for batch size = 16. GradInversion fails completely as in the above two rows; our method restores the image content, but is affected by images with the same label as in the middle two rows; the last two rows are ground truth.

\begin{table}
\centering
\caption{Compare the images recovered with our approach and GradInversion when there are duplicate labels in the batch. The former is GradInversion, the latter is our approach.}
\label{tab:compare}
\resizebox{\columnwidth}{!}{
\begin{tabular}{cccccc}
\toprule
Label Distribution       & MSE $\downarrow $    & SSIM $\downarrow$    & PSNR $\uparrow$ & LPIPS $\downarrow$   \\ 
\midrule
\textit{Repetition} 1         & 0.199 /0.018  &  0.453/  0.703& 10.639 / 19.122  &  0.633 /0.344    \\ \hline
\textit{Repetition} 2         & 0.334 / 0.031  & 0.227 /0.663 &5.227/18.258  &     0.820 / 0.392   \\ \hline
\textit{Repetition} 3         & 0.425/ 0.036  & 0.134/  0.606 & 4.150 /17.052 &  0.866    / 0.444    \\ \hline
\textit{Rep Amount} 2         & 0.458 / 0.050 & 0.044 / 0.548 & 3.893   / 13.626 &  0.896  /0.509 \\ \hline
\textit{Rep Amount} 4        & 0.464/ 0.068 & 0.021 / 0.461 &  3.953 /  12.164  &     0.890  /0.602  \\ \hline
\end{tabular}%
}
\end{table}

\begin{figure}
    \centering
    \includegraphics[width=0.4\textwidth]{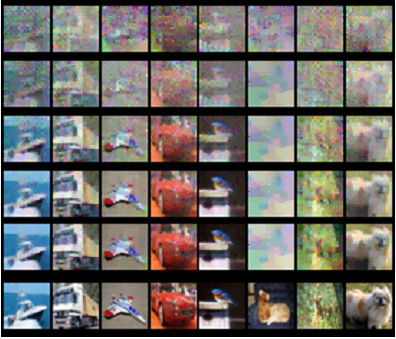}
    \caption{Batch image restoration from weights on CIFAR-10 with batch size of 8 with no duplicate label.
    The first to fifth lines are the results of local epochs equal to 1,2,4,8,16 respectively, and the last line is the ground truth.}
    \label{fig:weight_restore}
\end{figure}

\paragraph{Image Reconstruction from Weights}   Figure~\ref{fig:weight_restore} illustrates the results of recovering unique images in CIFAR-10 after training for different local epochs, supposing the batch size equals the number of local images. Our method can restore training data when the batch size and local epochs are both greater than 1.

\section{Conclusion}
\label{sec:conclusion}
In this article, we conduct privacy attacks on the federated learning scenarios of shared gradient (FedSGD) and shared model (FedAvg), respectively. The experimental results show the fragile side of the naive federated learning system when sharing parameters. Our method solves the limitation of not having the same labels in one batch and can recover batch images and labels from model parameters after multiple local training epochs. We also empirically proved that multiple model-gradient pairs could improve image recovery. Our experiments remind people not to rely solely on the privacy of the federation learning mechanism, and additional privacy-preserving techniques need to be designed for federation learning.

\appendix
\label{sec:reference_examples}
\bibliography{aaai22}
\end{document}